\documentclass[conference]{IEEEtran}
\IEEEoverridecommandlockouts
\usepackage{cite}
\usepackage{amsmath,amssymb,amsfonts}
\usepackage{algorithmic}
\usepackage{graphicx}
\usepackage{textcomp}
\usepackage{xcolor}
\def\BibTeX{{\rm B\kern-.05em{\sc i\kern-.025em b}\kern-.08em
    T\kern-.1667em\lower.7ex\hbox{E}\kern-.125emX}}

\begin{document}

\title{Gap the (Theory of) Mind: Sharing Beliefs About Teammates' Goals Boosts Collaboration Perception, Not Performance}
\author{
\IEEEauthorblockN{1\textsuperscript{st} Yotam Amitai}
\IEEEauthorblockA{\textit{Faculty of Data \& Decisions Science} \\
\textit{Technion - Israel Institute of Technology}\\
Haifa, Israel \\
0000-0002-0084-9739}
\and
\IEEEauthorblockN{2\textsuperscript{nd} Reuth Mirsky}
\IEEEauthorblockA{\textit{Department of Computer Science} \\
\textit{Tufts University}\\
Medford, Massachusetts, United States \\
0000-0003-1392-9444}
\and
\IEEEauthorblockN{3\textsuperscript{rd} Ofra Amir}
\IEEEauthorblockA{\textit{Faculty of Data \& Decisions Science} \\
\textit{Technion - Israel Institute of Technology}\\
Haifa, Israel \\
0000-0003-2303-3684}
}

\maketitle

\begin{abstract}
In human-agent teams, openly sharing goals is often assumed to enhance planning, collaboration, and effectiveness. However, direct communication of these goals is not always feasible, requiring teammates to infer their partner’s intentions through actions. Building on this, we investigate whether an AI agent’s ability to share its inferred understanding of a human teammate’s goals can improve task performance and perceived collaboration. Through an experiment comparing three conditions—no recognition (NR), viable goals (VG), and viable goals on-demand (VGod)—we find that while goal-sharing information did not yield significant improvements in task performance or overall satisfaction scores, thematic analysis suggests that it supported strategic adaptations and subjective perceptions of collaboration. Cognitive load assessments revealed no additional burden across conditions, highlighting the challenge of balancing informativeness and simplicity in human-agent interactions. 
These findings highlight the nuanced trade-off of goal-sharing: while it fosters trust and enhances perceived collaboration, it can occasionally hinder objective performance gains.
\end{abstract}


\section{Introduction}
In human-agent collaboration, effective teamwork often depends on the agent’s ability to interpret and act upon the human teammate’s intentions. Ad-hoc teamwork \cite{stone2010ad}, where team members must collaborate effectively without prior planning, exemplifies contexts where this capability is critical. Explainable AI (XAI) aims to address this by enhancing transparency and interpretability in AI systems, fostering shared mental models, trust, and mutual understanding \cite{harbers2012explanation, sridharan2019towards}.
The principles of goal-setting theory \cite{van2018goal} highlight the role of clear objectives in enhancing team performance by activating motivational drivers. Similarly, the concept of shared mental models \cite{kozlowski2006enhancing} underscores the benefits of a common understanding of tasks and objectives for improving coordination and effectiveness. Therefore, in decision-making settings or ad-hoc teamwork scenarios, the practice of sharing teammates' inferred beliefs about objectives is seen as crucial for fostering stronger collaborative alignment and ensuring all members are effectively oriented towards common goals. However, while transparency and information sharing are assumed to benefit collaboration, empirical studies increasingly suggest a complex relationship between these factors and actual performance outcomes in human-agent settings \cite{perez2023experimental, le2023trusting}. 

This study investigates the impact of varying levels of goal-sharing information on both subjective and objective metrics in human-agent collaboration. Specifically, we designed an experiment in a modified tool-fetching domain, where participants, acting as the ``worker'', were paired with an AI agent, the ``fetcher'', whose task was to infer and respond to the worker’s intended goal location. Three experimental conditions were tested: no recognition (NR), where participants received no access to the fetcher's beliefs about their goal; viable goals (VG), where participants continuously saw all goals the agent deemed viable based on the participant's actions; and viable goals on-demand (VGod), where participants could selectively view this information at their discretion. The study assesses whether access to the agent's mental model and how it perceives its teammate's goal would lead to measurable performance improvements or enhance perceived collaboration quality.

The contributions of this study are threefold. First, through a large-scale user study, we provide empirical evidence highlighting the limitations of information sharing in human-agent teamwork, specifically underscoring the disconnect between subjective satisfaction and objective performance outcomes.
Second, by examining three distinct conditions, our study exemplifies how varying access to an agent's inferred goals influences both the user experience and performance outcomes. Lastly, informed by related work and our empirical evaluation, we identify potential failure modes—such as cognitive overload and interpretation challenges—that highlight the need for future research to clarify the conditions under which goal-sharing truly benefits collaboration. 

\section{Related Work}

In this section, we review the existing literature on the effects of information sharing, explanations in human-agent collaboration, and cognitive load in teamwork settings. The aim is to contextualize the research on whether sharing thoughts about a teammate's goals improves objective performance and perceived collaboration.

\subsection{Information Sharing in Teamwork and Decision-Making}

Several studies have examined the impact of sharing information between teammates, particularly in decision-making contexts. Research has shown that contrary to intuition, presenting a player's thoughts or intentions to another teammate does not necessarily improve performance and can sometimes hinder it. For example, Pérez-D’Arpino et al. \cite{perez2023experimental} demonstrated that while intention-sharing in human-robot collaboration increased trust and perceived collaboration quality, it did not yield a significant improvement in task performance. Similarly, Le Guillou et al. \cite{le2023trusting} explored how AI agents' intention-communication raised trust levels but without any corresponding enhancement in objective performance.
The discrepancy between perceived and measured performance is not limited to information sharing but also prevails in information gathering: a recent human-robot interaction study showed that when a robot inquired about a human's intentions, the interaction was perceived as an interruption even when no objective interference performance was measured \cite{mannem2023exploring}.

These studies suggest that although sharing information might improve the subjective experience of collaboration, it does not always translate into better decision-making or task efficiency, particularly when real-time responses are required.

Our study provides further evidence that more information sharing does not necessarily improve teamwork. We identify critical failure modes that may underlie this phenomenon, helping to pinpoint conditions under which goal-sharing is genuinely beneficial.

\subsection{Explanations in Human-Agent Collaboration}

Explainable AI (XAI) has been a growing area of interest, particularly in the context of human-agent collaboration. The assumption is that providing transparent reasoning for an AI agent’s decisions helps human teammates align their goals and actions more effectively. Harbers et al. \cite{harbers2012explanation} proposed that explanations are critical for increasing shared mental models within human-agent teams. However, despite theoretical support, empirical studies often present mixed results. Several studies emphasize that while explanations can increase human trust in agents, the actual performance gains from explanations remain inconsistent \cite{sridharan2019towards, wagner2021explanation}.

A key challenge with explanations in ad-hoc teamwork is their potential to overload human team members with information. Cognitive load theory, introduced by Sweller \cite{sweller1988cognitive}, suggests that when users are presented with more information than they can effectively process, their decision-making suffers. This is particularly relevant in high-stakes, time-sensitive environments, where too much explanation can hinder quick, effective responses.

\subsection{Cognitive Load and Its Effects on Performance}

Cognitive load theory posits that human cognitive capacity is limited, and overloading this capacity can reduce the effectiveness of explanations or additional information. In the context of human-agent teamwork, explanations or shared information must strike a delicate balance between being informative and overwhelming. Studies such as Gajos and Chauncey \cite{gajos2017influence} explored how individual differences, such as personality traits and the need for cognition, impact users' ability to process AI-generated explanations. Their results show that users with high cognitive demands are more likely to feel overwhelmed by complex information, leading to decreased task performance.

Similarly, Nimmo et al. \cite{nimmo2024user} highlighted that explanations tailored to the individual's cognitive capacity may improve engagement and trust but do not necessarily lead to better decision-making or outcomes. This underscores the potential pitfall of over-relying on shared mental models in teamwork contexts, where increased cognitive load from additional information can detract from objective performance.

\subsection{Distinction Between Perceived and Objective Help}

The distinction between the subjective perception of help and objective performance improvement has been highlighted in multiple studies. Conati et al. \cite{conati2021toward} found that while AI-generated explanations increased perceived usefulness and trust in intelligent tutoring systems, they did not always lead to improved learning outcomes. Similarly, Millecamp et al. \cite{millecamp2019explain} discovered that users who perceived explanations as helpful were not necessarily more accurate in their decisions. This gap between perceived benefit and actual performance improvement is crucial for understanding why sharing thoughts may feel beneficial without objectively aiding teamwork.


\section{Experimental Methodology}
This section describes the experimental setup, the metrics collected, and the factors considered that might influence the results. Our study aimed to evaluate both subjective and objective outcomes when agents share their beliefs about their human teammate's goals in ad-hoc teamwork settings.

\subsection{Domain}
\begin{figure}
\centering
\includegraphics[width=1\linewidth]{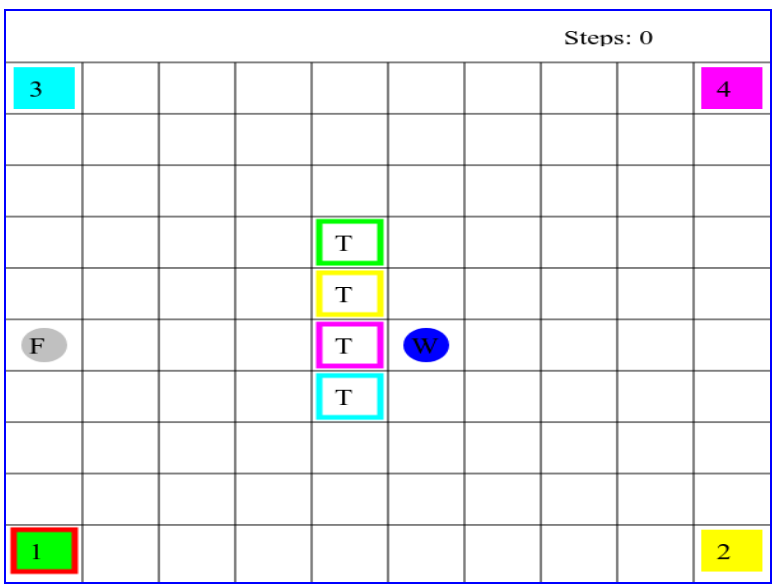}
\caption{The Worker-Fetcher Environment. The worker (W) moves to a numbered goal station (1, 2, 3, 4). The fetcher (F) infers the goal, retrieves the corresponding tool (T), and delivers it to the worker.}
\label{fig:domain}
\end{figure}

The environment used to evaluate our studies was a modified version of the \textit{tool fetching domain}\cite{macke2021expected,mirsky2020penny,suriadinata2021reasoning}. The environment, shown in Fig.\ref{fig:domain}, is a discrete-action collaborative game setup utilized to explore human-agent teamwork dynamics. The game involves two players: the \textit{worker} and the \textit{fetcher}. The worker, represented by a circular blue agent labeled $W$, is controlled by the human player. The primary objective of the worker is to navigate to a predetermined numbered station, denoted by a red frame, in a manner that is efficient and straightforward for the fetcher to interpret. The fetcher, an AI agent, does not initially know the worker's goal and must infer the correct station based on the worker's movements. The fetcher's goal is to retrieve a tool from the corresponding colored toolbox (labeled $T$) and deliver it to the worker at the goal station. The game's score is evaluated based on the number of game steps taken until the fetcher joins the worker at the goal. The game controls include basic directional movements (non-diagonal) and specific actions, such as working at the goal station and waiting, ensuring a straightforward yet strategic interaction environment for studying human-agent collaboration.

\subsection{Goal Recognition and Interaction}
The fetcher (agent) perceives the worker's goal using an inference method that decreases the probability of a station being the goal if an action is taken in an opposite direction. This is done by multiplying the current probability of the station by some learning rate parameter $\eta$ (close to 0) and then normalizing all station probabilities. This approach ensures that no station probability ever reaches 0, enabling inference even when the worker takes suboptimal actions. When a station probability exceeds some threshold $h$, the fetcher will assume this is the goal station and start progressing accordingly. Additionally, if multiple stations have equal probabilities (equal to the highest available probability) the fetcher will seek to advance toward common ground between these if possible. If no such action exists, the fetcher waits for more information. In our experiments, we set $\eta=0.05$ and $h = 1-\eta$.

\section{Experiment Design}
The experiment was designed to explore the effects of an agent sharing its understanding of the human player's goal during a collaborative teamwork task. Specifically, the human player could view the agent's interpretation of the player's intended goal. The agent would update its beliefs through a goal recognition technique based on the player's actions. This research was approved by the \textit{Anonymized Institute} Ethics Committee.

Participants were split into different conditions: those who received access to their agent teammate's thoughts and those who did not. We compared both subjective and objective outcomes across these groups. The specific conditions were:
\begin{itemize}
    \item \textbf{NR (No Recognition)}: Participants did not receive any access to the agent's goal recognition beliefs.
    \item \textbf{VG (Viable Goals)}: Participants could see all goals the agent perceived were viable based on the player's actions at all times.
    \item \textbf{VGod (Viable Goals On-Demand)}: Participants could view the agent's perceived viable goals when requested by clicking a button.
\end{itemize}

\subsection{Participants}
We recruited $N=313$ participants through Prolific\footnote{www.prolific.com}. After filtering for incomplete data, attention check failures, and outliers, $279$ participants remained in the final dataset. Participants were from English-speaking countries (e.g., UK, US, Canada, Australia), with an average age of 36 and a roughly equal gender distribution.

\subsection{Task and Procedure}
Participants were introduced to the worker-fetcher environment, where the player (the worker) navigated to a goal station, and the agent (the fetcher) inferred the worker's goal and retrieved the appropriate tool. The fetcher's ability to correctly and quickly infer the worker’s goal was the primary task, aiming to minimize the number of game steps. The path chosen by the player closely shaped the agent's ability to adapt to a new goal, as the agent dynamically refined its understanding in response to the player’s actions.

The experiment was divided into multiple stages:
\begin{enumerate}
    \item Participants first completed a tutorial to familiarize themselves with the task.
    \item Participants were asked to answer questions to test their attention and task comprehension in order to proceed.
    \item Participants were then placed into one of the three experimental conditions.
    \item In the experimental phase, the participant controlled the worker and navigated it toward their intended goal, while the fetcher had to infer the goal based on the worker's movements. In both VG and VGod conditions, participants could view the fetcher agent's perceived worker goal, while participants in the NR condition could not access such information. All participants played through four different game scenarios with rising difficulty. Each scenario was played three times to allow participants to consider different paths.
    \item Upon completion, participants completed questionnaires measuring subjective satisfaction and cognitive load.
\end{enumerate}

\subsection{Metrics Collected}
We measured objective performance metrics derived from the number of game steps and duration of participants in each scenario along with subjective self-reported assessments and ratings using 7-point Likert scales:
\begin{itemize}
    \item Cognitive load, using NASA Task Load Index ~\cite{hart1986nasa}.\\
    Example question: \textit{``How hurried or rushed was the pace of the task?''}
    \item Explanation satisfaction using Hoffman et al. ~\cite{hoffman2018metrics}.\\
    Example question: \textit{``Paying attention to the fetcher's intent improved my score.''}
    
    \item Need for cognition, using de Holanda Coelho et al. ~\cite{lins2020very}.\\ Example question: \textit{``Thinking is not my idea of fun.''}
    \item Attitude towards AI (AIAS-4) \cite{grassini2023development}. \\
    Example question: \textit{``I think AI technology is positive for humanity.''}
\end{itemize}
Additionally, open-ended feedback was collected for thematic analysis (e.g., \textit{``Please describe the fetcher's strategy as best you can.''}).
The full survey is available in the supplementary material. 

\subsection{Data Analysis}
Data pre-processing involved removing outliers, duplicates, and zero-variance features. The primary analysis compared performance and satisfaction metrics across the three experimental conditions. We further examined variations based on participants' need for cognition and attitudes toward AI. Statistical significance was determined through ANOVA.

To gain deeper insights into the factors influencing performance, we employed several analytical methods, including SHAP values in XGBoost models to pinpoint the most influential features, latent profile analysis to identify distinct participant clusters, and thematic analysis of textual responses to capture nuanced differences across conditions.

\section{Results and Experiment Discussion}

This section presents the results from our experiment, focusing on both objective performance and subjective perceptions across the three conditions: NR, VG, and VGod.

\subsection{Objective Performance}
Our primary objective performance metric was the number of steps taken to complete the task, which reflects both the efficiency of the player's path and its clarity, which aids the agent in quickly recognizing the player's goal. 

\begin{figure}
\centering
\includegraphics[width=1\linewidth]{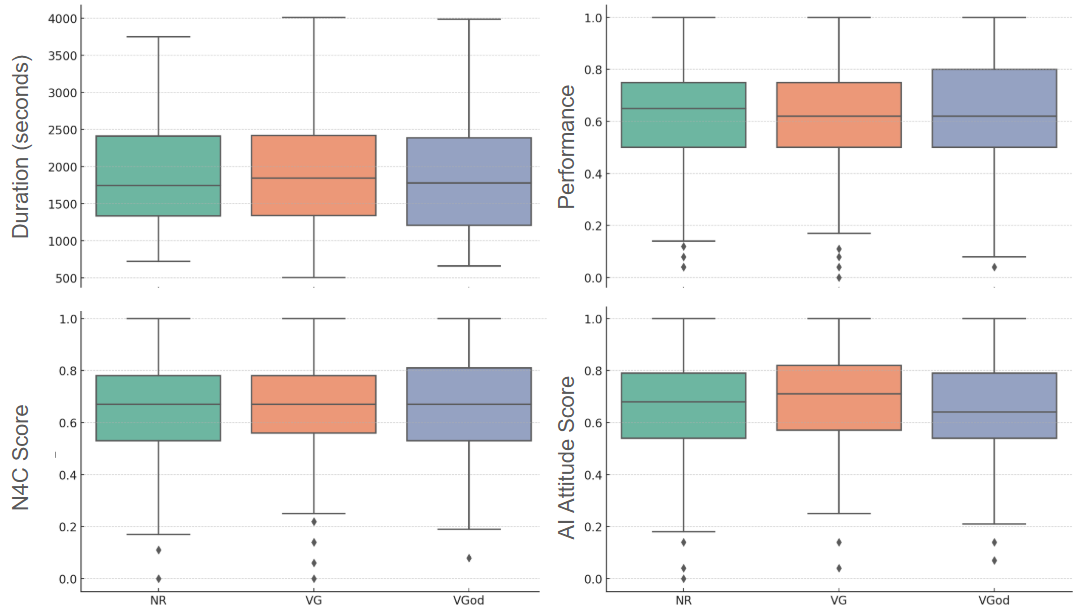}
\caption{Box plot distributions of task-related metrics across the experimental conditions: Duration (top-left), Performance (top-right), Need for Cognition (N4C) Score (bottom-left), and AI Attitude Score (bottom-right).}
\label{fig:performance_comparison}
\end{figure}

Figure \ref{fig:performance_comparison} indicates no significant improvement in task performance was observed for participants with access to the fetcher agent's perceived goals (VG and VGod) compared to those without this information. In addition to the number of steps, we compared task completion durations to assess whether access to the agent's inferred goals affected the speed of task execution. However, no significant differences were observed across conditions, suggesting that access to this information did not consistently translate into faster task completion.
As a countermeasure, we verified the distribution of participants' Need for Cognition (N4C) scores and attitudes toward AI across conditions and found no notable differences in these secondary measures.
ANOVA tests confirmed that there were no statistically significant differences between conditions: Performance- $F =  0.036$, $p = 0.965$, Duration- $F = 0.075$, $p = 0.928$, N4C Score- $F = 0.411$, $p = 0.663$, AI Attitude- $F =  2.404$, $p = 0.092$. This absence of significant differences implies that the additional information given to participants in the VG and VGod conditions \textbf{did not result in objectively enhanced efficiency}, nor did it reveal any differences in traits or characteristics between the groups.

\subsection{Subjective Perception}
We conducted a one-way ANOVA to assess the effect of explanation type on participants' overall satisfaction scores. The analysis revealed no statistically significant differences between the conditions ($F = 2.24, p = 0.11$), indicating that the type of explanation provided did not have a meaningful impact on participants' reported satisfaction.

\paragraph{Effect of Choice on Perceived Helpfulness}
To further investigate the role of information access, we examined the frequency with which VGod participants requested to view additional information and its relationship with satisfaction and performance metrics. The analysis found no significant correlation between the frequency of information requests and either satisfaction or performance outcomes.

Interestingly, a participant in one of the pilot think-aloud interview studies consciously chose not to press the button to access information, preferring to reserve it for moments of necessity. This highlights how individual strategies for utilizing on-demand information may vary, even when the option is available at no cost, further supporting the notion that \textbf{access, rather than usage frequency, is what influences perceived helpfulness}.

\paragraph{Thematic Analysis of Open-Ended Responses}
Qualitative analysis of open-ended responses provided additional insights. The thematic analysis was conducted using ChatGPT-4o with a structured prompt to ensure a consistent and comprehensive evaluation of participant responses. The prompt instructed the LLM to identify themes, patterns, insights and differences in participant responses between conditions and to provide examples.
The analysis provided by ChatGPT was carefully reviewed and validated by the first author to ensure the reliability and accuracy of the identified themes.
\\
\emph{Participant strategy.} Participants prioritized path clarity, directness, and effective fetcher guidance across all conditions. However, $40\%$ of Condition NR participants emphasized early signaling to the fetcher (e.g., ``I tried to move in a way that signaled as early as possible which station I was heading to''), while $45\%$ of those in Condition VG focused on minimizing the fetcher’s steps relative to their own (e.g., ``I tried to minimize the worker’s steps while reducing the fetcher’s steps at the same time''). In contrast, $50\%$ of Condition VGod participants relied on trial-and-error strategies to adapt to fetcher behavior (e.g., ``I tried different paths and used the fetcher’s response to refine my movements''). 
\\
\emph{Fetcher strategy.}
Across the conditions, $65\%$ of participants consistently noted that the fetcher’s movements were based on the worker's direction, with minor variations between conditions. In the NR condition, $40\%$ observed randomness or inefficiency (e.g., ``It seemed to follow my direction but take the longest route.''). In the VG condition, $50\%$ described the fetcher waiting to build confidence (e.g., ``It only moved once boxes in opposite directions had been ruled out.''). In the VGod condition, $25\%$ highlighted stronger interaction dynamics, such as, ``I tried to move in ways that allowed the fetcher to minimize steps.'' 
\\
\emph{Retrospection.} 
Participants in the NR condition frequently reported difficulty understanding the fetcher's behavior, with $36\%$ expressing confusion and perceiving some actions as random (e.g., ``Not really, it went in the direction I wanted but took a longer route''). Conversely, VG participants demonstrated a higher tendency to adjust their strategies proactively, with $49\%$ mentioning deliberate efforts to influence the fetcher (e.g., ``I would try to coax the fetcher to move towards the toolbox I wanted''). The VGod condition showed mixed results, with $40\%$ highlighting their understanding of the fetcher’s logic in ambiguous scenarios, but $45\%$ still reporting inefficiencies or lack of responsiveness (e.g., ``It seemed like it became more inefficient, especially as it brought the tool back''). These trends suggest that \textbf{enhanced visual guidance in VG and VGod conditions aids strategic behavior but does not eliminate all confusion}.

While no significant differences were found in satisfaction scores across conditions, thematic analysis revealed that participants in VG and VGod conditions adopted more strategic behaviors, such as minimizing the fetcher’s steps or using trial-and-error to align with the agent. These findings suggest that goal-sharing information, while not reflected in performance metrics or satisfaction scales, provided subjective value by enabling participants to better understand and adapt to the agent’s behavior, enhancing the perceived quality of collaboration.

\subsection{Cognitive Load}
\begin{figure}
\centering
\includegraphics[width=1\linewidth]{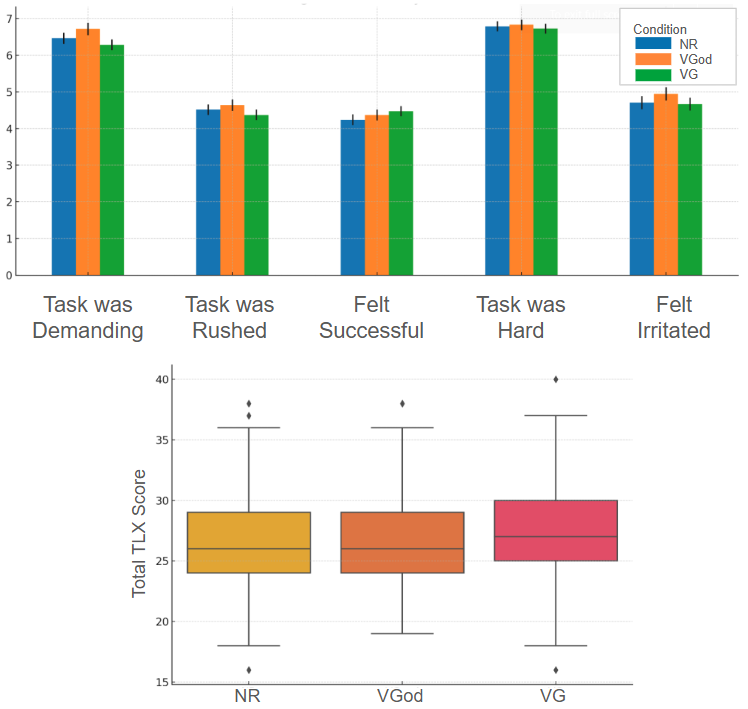}
\caption{NASA-TLX scores across conditions, adjusted s.t. higher scores indicate better outcomes.}
\label{fig:tlx}
\end{figure}

To understand the impact of cognitive load, we analyzed responses to the NASA-TLX scale\cite{hart1986nasa} across conditions as reported in Figure \ref{fig:tlx}. An ANOVA conducted on the aggregated scores revealed no statistically significant differences in overall TLX scores across conditions ($F = 1.455, p=0.235$), suggesting that access to the agent's perceived goals did not impose an additional cognitive burden on participants.
This result aligns well with the results of the objective performance (Figure \ref{fig:performance_comparison}), showing a consistent outcome in terms of the cost of processing the additional information.

\subsection{Demographics}
We analyzed demographic data to determine whether task familiarity or participants' backgrounds, particularly their experience with AI, influenced the results. Our analysis revealed no significant correlations between demographic factors, including AI experience, and either performance or satisfaction metrics, indicating that these variables did not meaningfully impact outcomes.

\section{Discussion}
Our study examined the impact of sharing an agent’s inferred goals with a human teammate in ad-hoc teamwork settings. The study revealed a disconnect between subjective perceptions of collaboration and objective performance in human-agent teamwork. Sharing the agent’s inferred goals did not significantly improve task performance or overall satisfaction across the NR, VG, and VGod conditions. Measures such as game steps, task completion duration, and cognitive load showed no significant differences between conditions.

Qualitative analysis highlighted varied strategies: NR participants often reported confusion, while VG and VGod participants adjusted their actions to align with the agent’s behavior. However, these adaptations did not translate into measurable performance gains. Additionally, demographic factors like AI experience or task familiarity had no meaningful impact on outcomes, underscoring the generalizability of the results.

These findings emphasize the challenge of translating transparency into performance benefits and highlight the importance of designing goal-sharing mechanisms that balance clarity and cognitive load effectively.

This section discusses potential reasons for this outcome, explores the role of subjective perception, considers cognitive load factors, and highlights implications for designing human-agent teams.

\subsection{Why Does Information Sharing Not Improve Performance?}
One possible explanation is the balancing effect of the increased cognitive overload. Cognitive load theory suggests that providing additional information, particularly when complex, can overwhelm users, reducing their ability to process and effectively use the information \cite{sweller1988cognitive}. While overall cognitive load measures showed no significant differences across conditions, we do not rule out the possibility of localized or task-specific moments of cognitive overload. For example, interpreting the agent's inferred goals in real-time may have caused brief spikes in cognitive demand in critical moments, impacting performance without significantly affecting average cognitive load scores. Misinterpreting these goals could also lead to inefficient strategies, further limiting performance.

As cognitive load in this study was self-reported, it may lack the granularity to capture subtle variations. A within-subject design, where participants experience and compare multiple conditions, along with more dynamic measures, could better reveal the interplay between information sharing, cognitive load, and task performance.

\subsection{The Role of Subjective Perception}
The results highlight that even if information sharing does not objectively improve task performance, it can enhance the human teammate’s experience. Participants reported feeling more in control and satisfied with the agent’s actions when they had access to the agent's perceived goals. This subjective improvement is important because positive user experience and satisfaction are valuable in human-agent teamwork, especially when trust and collaboration quality are crucial. These findings suggest that information sharing may contribute to building trust and a sense of partnership, even if it does not directly enhance efficiency.

\subsection{Cognitive Load Considerations}
The consistent cognitive load ratings across conditions suggest that access to the agent's inferred goals did not impose a substantial burden. However, specific participants may have been more affected by cognitive load based on individual differences such as cognitive ability, familiarity with similar tasks, and personal preferences for detailed information. A higher cognitive load might obscure the benefits of information sharing by diminishing users' capacity to fully leverage the agent’s insights. This factor should be carefully considered in future designs, as balancing informativeness and simplicity is essential for effective explanations.

\subsection{Implications for Human-Agent Teamwork}
These findings offer valuable insights for designing human-agent teams where user perception of the collaboration is as important as performance outcomes. In settings where trust, satisfaction, and perceived support are critical, providing explanations or sharing the agent’s inferred goals could improve collaboration quality even if it does not enhance task efficiency. Future designs should consider how to present information in ways that are intuitive and minimize cognitive load, potentially by adapting explanations based on user preferences or characteristics.

\subsection{Failure Modes in Evaluating Explanations}
Evaluating explanations in ad-hoc teamwork requires careful consideration of potential flaws in study design and evaluation methods, as these can compromise the validity and generalizability of findings. We outline potential \textbf{failure modes}, including those we have addressed and others we were unable to fully mitigate, which may distort the evaluation process.
Addressing these failure modes is critical to improving the study and evaluation of explanations.

\paragraph{Task design flaws}
Ambiguous instructions can lead to varied interpretations and unreliable data, while insufficient motivation may affect participants’ engagement and performance, as highlighted in prior research linking motivation to task success \cite{steinmayr2019importance}. In this study, we addressed task design flaws by running several pilot studies, including a think-aloud study, refining instructions, and providing clear tutorials and performance-based incentives to ensure participant understanding and motivation. 

\paragraph{Explanation quality flaws}
Cognitive overload, where the information exceeds a user’s processing capacity, can reduce comprehension and usability \cite{sweller1988cognitive}. Similarly, unhelpful or irrelevant explanations fail to address users’ needs, limiting their utility \cite{gregor1999explanations}. Poor user interface design can further hinder the effectiveness of explanations, as it impacts users’ ability to understand and engage with the system \cite{elshan2022understanding}.
We addressed explanation quality flaws by designing explanations to balance informativeness and simplicity, ensuring they were concise and relevant to participants’ tasks. Additionally, the user interface and instructions were iteratively refined through pilot studies to enhance clarity and usability, minimizing cognitive overload and maximizing engagement.

\paragraph{Explainee variability}
Individual differences, such as personality traits, cognitive abilities, and prior experience, influence how explanations are perceived and utilized \cite{conati2021toward}. Trust in AI systems, shaped by factors like transparency and reliability, can affect how users interact with explanations \cite{hoff2015trust}. Additionally, automation bias, where users over-rely on AI systems, may undermine their critical engagement with explanations \cite{parasuraman1997humans}. Encouraging user agency can help mitigate these effects by fostering active involvement and critical evaluation of AI outputs \cite{buccinca2021trust}.
We addressed explainee variability by collecting demographic and background data to account for differences in personality, cognitive abilities, and experience. Additionally, we provided options for user control to promote agency and reduce automation bias, encouraging active engagement with the system.

\paragraph{Contextual and environmental factors}
Environmental distractions during remote studies can introduce noise, while misalignment between the study setup and real-world contexts may reduce the applicability of findings \cite{nielsen1994usability}.
These factors were not controlled or addressed in this study.

\paragraph{Measurement and evaluation flaws} Using invalid or inconsistent metrics makes it challenging to assess the impact of explanations and compare results across studies. Reliable, standardized evaluation methods are essential for producing robust and generalizable insights.
We addressed measurement and evaluation flaws by employing validated, standardized metrics for performance, satisfaction, and cognitive load, ensuring consistent and reliable evaluation.

\section{Conclusion}
This paper presented the results of a user study exploring the role of sharing perceived goals in human-AI teams. While sharing an AI agent’s inferred goals with human teammates does not necessarily improve task performance, it provides subjective value in enhancing the perceived quality of collaboration. Although no significant differences were found in satisfaction scores across conditions, the thematic analysis revealed that participants in VG and VGod conditions adopted more strategic behaviors, such as minimizing the fetcher’s steps or using trial-and-error to align with the agent. This result suggests that goal-sharing information, while not reflected in objective metrics, allowed participants to better understand and adapt to the agent’s behavior.

These results align with cognitive load theory, indicating that additional information may overwhelm users, reducing the potential benefits of transparency in real-time settings. Consequently, human-agent teamwork designs should carefully balance informativeness and simplicity, focusing on user needs and preferences to optimize collaboration outcomes.

This study highlights the nuanced impact of sharing an AI agent’s inferred goals on human-agent teamwork, where access to such information offers subjective benefits but does not consistently improve task performance. The absence of clear objective benefits from goal-sharing points to several possible failure modes. These findings underscore the need for a more comprehensive understanding of when and how additional information benefits human-agent collaboration. Current models of information sharing may inadequately address the diverse ways users process and act on agent-provided insights, particularly in real-time and high-stakes environments. As such, future work should explore the boundary conditions under which transparency enhances task effectiveness versus those where it can potentially hinder it. Developing robust frameworks for addressing failure modes will be critical to designing AI explanations that are both intuitive and effective in enhancing collaboration.

\section{Acknowledgments}
Funded by the European Union (ERC, Convey, 101078158). Views and opinions expressed are, however, those of the author(s) only and do not necessarily reflect those of the European Union or the European Research Council Executive Agency. Neither the European Union nor the granting authority can be held responsible for them. OA also acknowledges the support of the Schmidt Career Advancement Chair in Artificial Intelligence

\bibliographystyle{IEEEtran} 
\bibliography{main}

\end{document}